\documentclass[runningheads]{llncs}
\usepackage{graphicx}
\usepackage{amsmath,amssymb} 
\usepackage{booktabs}
\usepackage{color}

\begin{document}
\title{Coloring with Words: Guiding Image Colorization Through Text-based Palette Generation} 

\titlerunning{Coloring with Words}
%
\makeatletter
\renewcommand*{\@fnsymbol}[1]{\ifcase#1\or*\else\@arabic{\numexpr#1-1\relax}\fi}
\makeatother

\author{Hyojin Bahng\thanks{These authors contributed equally.}\inst{1}
\and Seungjoo Yoo\textsuperscript{*}\inst{1}
\and Wonwoong Cho\textsuperscript{*}\inst{1}
\and David Keetae Park\index{Park, David} \inst{1,3} 
\and Ziming Wu\inst{2}
\and Xiaojuan Ma\inst{2}
\and Jaegul Choo\inst{1,3}}
%
\authorrunning{Hyojin Bahng, Seungjoo Yoo, Wonwoong Cho}
%

\institute{Korea University \email{\{hjj552,seungjooyoo,tyflehd21,heykeetae,jchoo\}@korea.ac.kr}
\and
Hong Kong University of Science and Technology \\
\email{zwual@connect.ust.hk, mxj@cse.ust.hk}
\and
Clova AI Research, NAVER Corp.}
\maketitle              
\begin{abstract}
This paper proposes a novel approach to generate multiple color palettes that reflect the semantics of input text and then colorize a given grayscale image according to the generated color palette. In contrast to existing approaches, our model can understand rich text, whether it is a single word, a phrase, or a sentence, and generate multiple possible palettes from it. For this task, we introduce our manually curated dataset called Palette-and-Text (PAT). Our proposed model called Text2Colors consists of two conditional generative adversarial networks: the text-to-palette generation networks and the palette-based colorization networks. The former captures the semantics of the text input and produce relevant color palettes. The latter colorizes a grayscale image using the generated color palette. Our evaluation results show that people preferred our generated palettes over ground truth palettes and that our model can effectively reflect the given palette when colorizing an image.

\end{abstract}

\begin{figure}[t]
\centering
\includegraphics[width=\textwidth]{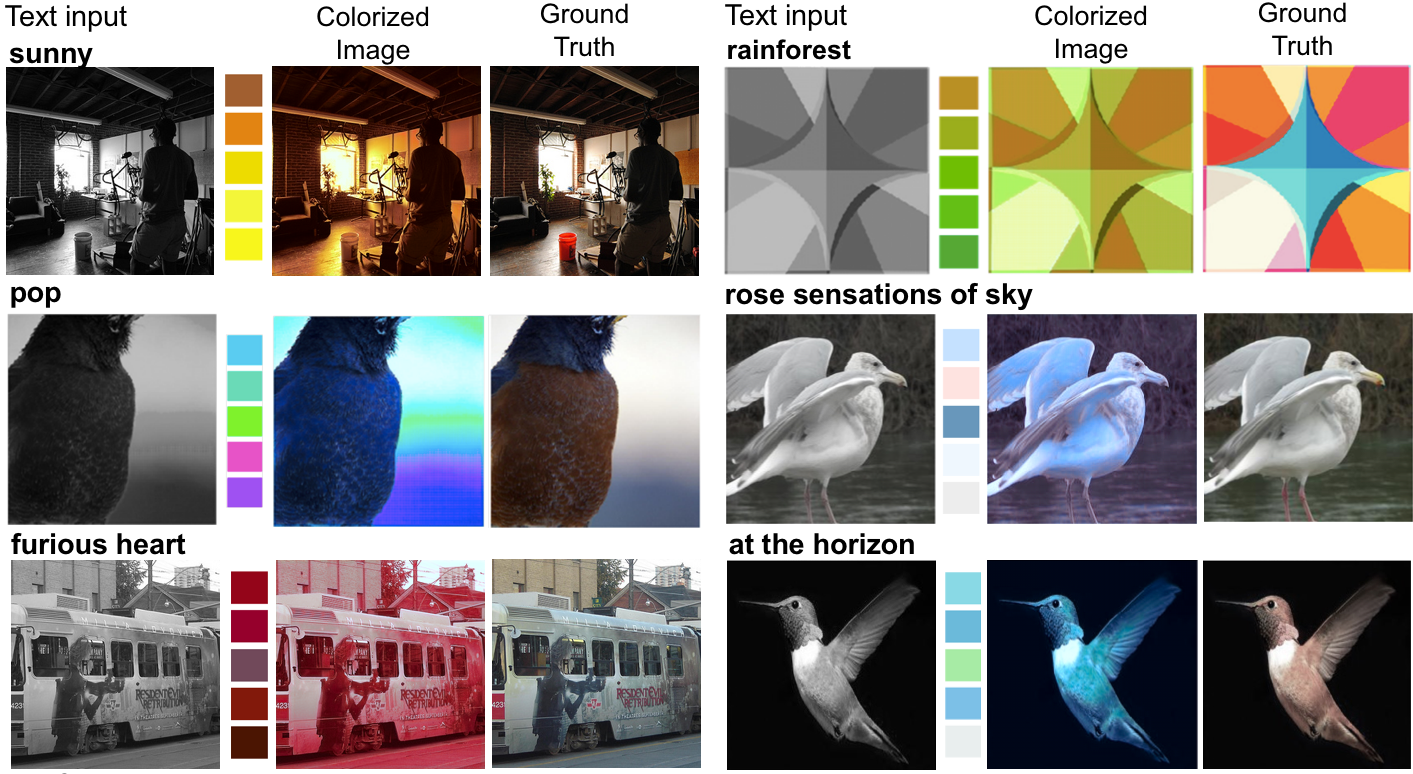}
\caption{\textbf{Colorization results of Text2Colors given text inputs.} The text input is shown above the input grayscale image, and the generated palettes are on the right of the grayscale image. The color palette is well-reflected in the colorized image when compared to the ground truth image. Our model is applicable to a wide variety of images ranging from photos to patterns (top right).}\label{fig:}
\end{figure}

\begin{figure}[t]
\centering
\includegraphics[width=\textwidth]{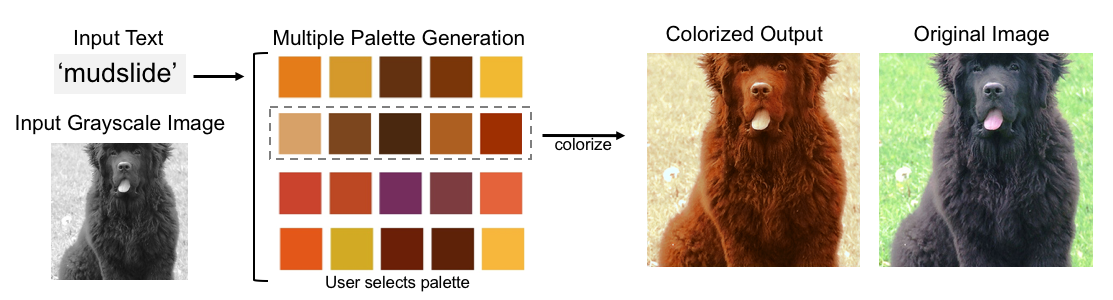}
\caption{\textbf{How Text2Colors works.} Our model can produce a diverse selection of palettes when given a text input. Users can optionally choose which palette to be applied to the final colorization output.}\label{fig:cover}
\end{figure}

\section{Introduction}
Humans can associate certain words with certain colors. The real question is, can machines effectively learn the relationship between color and text? Using text to express colors can allow ample room for creativity, and it would be useful to visualize the colors of a certain semantic concept. For instance, since colors can leave a strong impression on people~\cite{labrecque2012exciting}, corporations often decide upon the season`s color theme from marketing concepts such as `passion.' Through text input, even people without artistic backgrounds can easily create color palettes that convey high-level concepts. Since our model uses text to visualize aesthetic concepts, its range of future applications can encompass text to even speech. 

Previous methods have a limited range of applications as they only take a single word as input and can recommend only a single color or a color palette in pre-existing datasets~\cite{heer2012color,chuang2008probabilistic,kawakami2016character,monroe2017colors}. Other studies have further attempted to link a single word with a multi-color palette~\cite{liu2014autostyle,solli2010color} since multi-color palettes are highly expressive in conveying semantics~\cite{kobayashi2009color}. Compared to these previous studies, our model can generate multiple plausible color palettes when given rich text input, including both single- and multi-word descriptions, greatly increasing the boundary of creative expression through words.

In this paper, we propose a novel method to generate multiple color palettes that convey the semantics of rich text and then colorize a given grayscale image according to the generated color palette. Perception of color is inherently multimodal~\cite{charpiat2008automatic}, meaning that a particular text input can be mapped to multiple possible color palettes. To incorporate such multimodality into our model, our palette generation networks are designed to generate multiple palettes from a single text input. We further apply our generated color palette to the colorization task. Motivated from previous user-guided colorizations that utilize color hints given by users~\cite{xiao2018interactive,zhang2017real}, we design our colorization networks to utilize color palettes during the colorization process. Our evaluation demonstrates that the colorized outputs do not only reflect the colors in the palette but also convey the semantics of the text input.

The contribution of this paper includes: 
 
\noindent (1) We propose a novel deep neural network architecture that can generate multiple color palettes based on natural-language text input. 

\noindent (2) Our model is able to use the generated palette to produce plausible colorizations of a grayscale image. 

\noindent (3) We introduce our manually curated dataset called Palette-and-Text (PAT), which includes 10,183 pairs of a multi-word text and a multi-color palette.~\footnote{Dataset and codes are publicly available at https://github.com/awesome-davian/Text2Colors/}

\section{Related Work}
\subsubsection{Color Semantics}
Meanings associated with a color are both innate and learned~\cite{crozier1996psychology}. For instance, red can make us instinctively feel alert~\cite{crozier1996psychology}. Since color has a strong association with high-level semantic concepts~\cite{de2001colours}, producing palettes from text input is useful in aiding artists and designers~\cite{kobayashi2009color} and allows automatic colorization from palettes~\cite{xiao2018interactive,cho2017palettenet}. A downside to using text to choose a filter is that filter names do not usually convey the filter's colors~\cite{liu2014autostyle}, thus making it difficult for users to find the filter that matches their taste just by looking at filter names. To bridge this discrepancy between color palettes and their names, palette recommendation based on user text input has long been studied. Query-based methods~\cite{liu2014autostyle,solli2010color} use text inputs to query an image from an image dictionary where colors are extracted from the queried image to make an associated palette. This method is problematic in that the text input is mapped to the image content of the queried image rather than the color that the text implies. Instead of looking for a target directly, learning-based approaches~\cite{jahanian2017colors,murray2012toward,mcmahan2015bayesian} match or generate color palettes to their linguistic descriptions by learning their semantic association from large-scale data. However, our model is the only generative model that supports phrase-level text input.  

\subsubsection{Conditional GANs}
Conditional generative adversarial networks (cGAN) are GAN models that use conditional information for the discriminator and the generator~\cite{mirza2014conditional}. cGANs have drawn promising results for image generation from text~\cite{reed2016generative,reed2016learning,zhang2017stackgan} and image-to-image translation~\cite{kim2017learning,isola2017image,choi2017stargan}. StackGAN~\cite{zhang2017stackgan} is the first model to use conditional loss for text to image synthesis. Our model is the first to utilize the conditioning augmentation technique from StackGAN to output diverse palettes even when given the same input text.

\subsubsection{Interactive Colorization} 
Colorization is a multimodal task and desired colorization results for the same object may vary from person to person~\cite{charpiat2008automatic}. A number of studies introduce interactive methods that allow users to control the final colorization output~\cite{zhang2017real,li2015image}. In these models, users directly interact with the model by pinpointing where to color. Even though these methods achieve satisfactory results, a limitation is that users need to have a certain level of artistic skill. Thus instead of making the user directly color an image, other studies take a more indirect approach by utilizing color palettes to recolor an image~\cite{chang2015palette,cho2017palettenet}. Palette-based filters of our model are an effective way for non-experts to recolor an image~\cite{chang2015palette}. 

\subsubsection{Sequence-to-Sequence with Attention} Recurrent Neural Networks (RNNs) are a popular tool due to their superior ability to learn from sequential data. RNNs are used in various tasks including sentence classification~\cite{tang2015document}, text generation~\cite{sutskever2011generating}, and sequence-to-sequence prediction~\cite{sutskever2014sequence}. Incorporating attention into a sequence-to-sequence model is known to improve the model performance~\cite{luong2015effective} as networks learn to selectively focus on parts of a source sentence. This allows a model to learn relations between different modalities as is done by our model (e.g., text - colors, text - action~\cite{ahn2017text2action}, and English - French~\cite{vaswani2017attention}). 

\section{Palette-and-Text (PAT) Dataset}
\label{sec:PAT}
\begin{figure}[t]
\centering
\includegraphics[width=\textwidth]{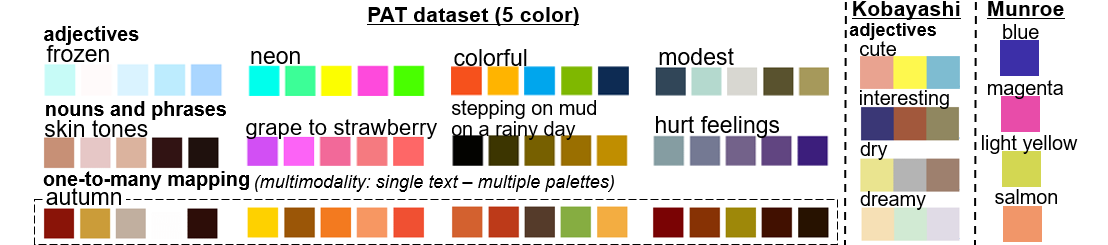}
\caption{\textbf{Our Palette-and-Text (PAT) dataset.}  On the left are diverse text-palette pairs included in PAT. PAT has a very wide range of expression, especially when compared to existing datasets. Our dataset is designed to address rich text and multimodality, where the same word can be mapped to a wide range of possible colors.}\label{fig:dataset_examples}
\end{figure}

This section introduces our manually curated dataset named Palette-and-Text (PAT). PAT contains 10,183 text and five-color palette pairs, where the set of five colors in a palette is associated with its corresponding text description as shown in Figs.~\ref{fig:dataset_examples}(b)-(d). Words vary with respect to their relationships with colors; some words are direct color words (e.g., pink, blue, etc.) while others evoke a particular set of colors (e.g., autumn or vibrant). To the best of our knowledge, there has been no dataset that matches a multi-word text and its corresponding 5-color palette. This dataset allows us to train our models for predicting semantically consistent color palettes with textual inputs. 

\subsubsection{Other Color Datasets} 
Munroe`s color survey~\cite{munroe2010color} is a widely used large-scale color corpus. Based on crowd-sourced user judgment, it matches a text to a single color. Another dataset, Kobayashi`s Color Image Scale~\cite{kobayashi2009color}, is a well-established multi-color dataset. Kobayashi only uses 180 adjectives to express 1170 three-color palettes, which greatly limits its range of expression. In contrast, our dataset is made up of 4,312 unique words. This includes much more text that was not traditionally used to express colors. Our task requires a more sophisticated dataset like PAT, that matches a text to multiple colors and is large enough for a deep learning model to learn from.
\subsubsection{Data Collection} 
We generated our PAT dataset by refining user-named palette data crawled from a community website called color-hex.com. Thousands of users upload custom-made color palettes on color-hex, and thus our dataset was able to incorporate a wide pool of opinions. We crawled 47,665 palette-text pairs and removed non-alphanumerical and non-English words. Among them, we found that users sometimes assign palette names in an arbitrary manner, missing their semantic consistency with their corresponding color palettes. Some names are a collection of random words (e.g., `mehmeh' and 
`i spilled tea all over my laptop rip'), or are riddled with typos (e.g., `cause iiiiii see right through you boyyyyy' and `greene gardn'). Thus, using unrefined raw palette names would hinder model performances significantly. 

To refine the noisy raw data, four annotators voted whether the text paired with the color palette properly matches its semantic meanings. We then used only the text-palette pairs in which at least three annotators out of four agreed that semantic matching exists between the text and color palette. Including text-palette pairs in the dataset only when all four annotators agree was found to be unnecessarily strict, leaving not much room for personal subjectivity. Annotator’s perception is inherently subjective, meaning that a text-palette pair perfectly plausible to one person may not be agreeable to another. We wanted to incorporate such subjectivity by allowing a diverse selection of text-palette pairs. Mis-spelling and punctuation errors were manually corrected after the annotators finished sorting out the data.

\begin{figure}[t]
\centering
\includegraphics[width=\textwidth]{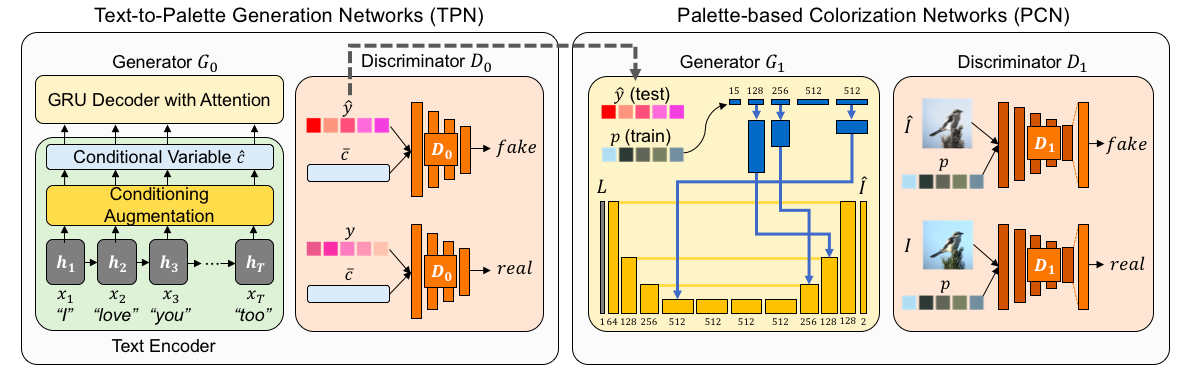}
\caption{\textbf{Overview of our Text2Colors architecture.} During training, generator $G_{0}$ learns to produce a color palette $\hat{y}$ given a set of conditional variables $\hat{c}$ processed from input text $x=\{x_{1},\cdots,x_{T}\}$. Generator $G_{1}$ learns to predict a colorized output of a grayscale image $L$ given a palette $p$ extracted from the ground truth image. At test time, the trained generators $G_{0}$ and $G_{1}$ are used to produce a color palette from given text and then colorize a grayscale image reflecting the generated palette.} \label{fig:full_model}
\end{figure}

\section{Text2Colors: Text-Driven Colorization}
Text2Colors consists of two networks: Text-to-Palette Generation Networks (TPN) and Palette-based Colorization Networks (PCN). We train the first networks to generate color palettes given a multi-word text and then train the second networks to predict reasonable colorizations given a grayscale image and the generated palettes. We utilize conditional GANs (cGAN) for both networks.

\subsection{Text-to-Palette Generation Networks (TPN)}
\label{sec:TPN}
\subsubsection{Objective Function}
\begin{figure}[t]
\centering
\includegraphics[width=\textwidth]{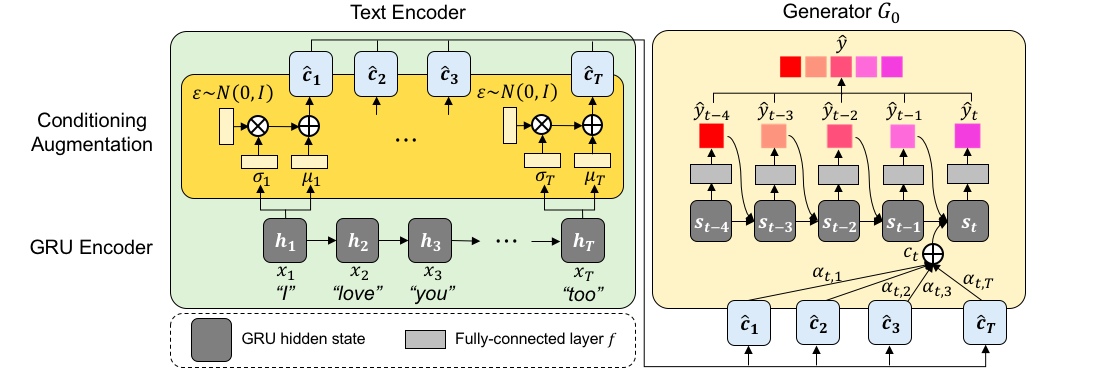}
\caption{Model architecture of a generator $G_{0}$ that produces the $t$-th color in the palette given a sequence of conditioning variables $\hat{c}=\{\hat{c}_{1},\cdots,\hat{c}_{T}\}$ processed from an input text $x=\{x_{1},\cdots,x_{T}\}$. Note that randomness is added to the encoded representation of text before it is passed to the generator.}\label{fig:ca_model}
\end{figure}

In this section, we illustrate the Text-to-Palette Generation Networks shown in Figs.~\ref{fig:full_model} and \ref{fig:ca_model}. TPN produces reasonable color palettes associated with the text input. Let $x_{i}\in\mathbb{R}^{300}$ be word vectors initialized by 300-dimensional pre-trained vectors from GloVe~\cite{pennington2014glove}.
Words not included in the pre-trained set are initialized
randomly. Using the CIE \emph{Lab} space for our task, $y\in\mathbb{R}^{15}$
represents a 15-dimensional color palette consisting of five colors with \emph{Lab}
values. After a GRU encoder encodes $x$ into hidden states $h=\{h_{1},\cdots,h_{T}\}$, we add random noise to the encoded representation of text by sampling latent variables $\hat{c}$ from a Gaussian distribution $\mathcal{N}(\mu(h),\Sigma(h))$. The sequence of conditioning vectors $\hat{c}=\{\hat{c}_{1},\cdots,\hat{c}_{T}\}$ is given as \textit{condition} for the generator to output a palette $\hat{y}$, while its mean vector $\bar{c}=\frac{1}{T}{\sum_{i=1}^{T}\hat{c}}$ is given as the condition for the discriminator. Our objective function of the first cGAN can be expressed as 
\begin{equation}
L_{D_{0}}=\mathbb{E_{\mathit{y\sim P_{data}}}}[\log D_{0}(\bar{c},y)]+\mathbb{E_{\mathrm{\mathit{x\sim P_{data}}}}\mathrm{[\log(1-\mathit{D_{0}}(\mathit{\bar{c},\hat{y}}))],}}
\end{equation}
\begin{equation}
\begin{aligned}
L_{G_{0}}=\mathbb{E_{\mathit{x\sim P_{data}}}}[\log(1-\mathit{D}_{0}(\mathit{\bar{c},\hat{y}}))],
\end{aligned}
\end{equation}
where discriminator ${D}_{0}$ tries to maximize $L_{D_{0}}$ against generator $G_{0}$ that tries to minimize $L_{G_{0}}$. The pre-trained word vectors $x$ and the real color palette $y$ is sampled from true data distribution $P_{data}$.

Previous approaches have benefited from mixing the GAN objective with $L_2$ distance~\cite{pathak2016context} or $L_1$ distance~\cite{isola2017image}. We have explored previous loss options and found the Huber (or smooth $L_1$) loss to be the most effective in increasing diversity among colors in generated palettes. The Huber loss is given by
\begin{equation}
\begin{aligned}
L_{H}(\hat{y},y)=\begin{cases}
\frac{1}{2}(\hat{y}-y)^{2}\;\;\mathrm{for}\;\;\left|\hat{y}-y\right|\leq\delta\\
\delta\left|\hat{y}-y\right|-\frac{1}{2}\delta^{2}\;\;\mathrm {otherwise}.
\end{cases}
\end{aligned}
\end{equation}
This loss term is added to the generator's objective function to force the generated palette to be close to the ground truth palette. We also adopted the Kullback-Leibler (KL) divergence regularization term~\cite{zhang2017stackgan}, i.e., 
\begin{equation}
D_{KL}(\mathcal{N}(\mu(h),\Sigma(h))\parallel\mathcal{N}(0,I)), 
\end{equation}
which is added to the generator's objective function to further enforce the smoothness over the conditioning manifold. Our final objective function is
\begin{equation}
L_{D_{0}}=\mathbb{E_{\mathit{y\sim P_{data}}}}[\log D_{0}(\bar{c},y)]+\mathbb{E_{\mathrm{\mathit{x\sim P_{data}}}}\mathrm{[\log(1-\mathit{D_{0}}(\mathit{\bar{c},\hat{y}}))],}}
\end{equation}
\begin{equation}
\label{eq:TPN generator}
\begin{aligned}
L_{G_{0}}=\mathbb{E_{\mathit{x\sim P_{data}}}}[\log(1-\mathit{D}_{0}(\mathit{\bar{c},\hat{y}}))]+\lambda_{H}L_{H}(\hat{y},y)\\+\lambda_{KL}D_{KL}(\mathcal{N}(\mu(h),\Sigma(h))\parallel\mathcal{N}(0,I)),
\end{aligned}
\end{equation}
$\lambda_{H}$ and $\lambda_{KL}$ are the hyperparameters to balance the three terms in Eq.~\ref{eq:TPN generator}. We set $\delta=1,\lambda_{H}=100,\lambda_{KL}=0.5$ in our model.

\subsubsection{Networks Architecture}
\paragraph{Encoding Text through Conditioning Augmentation.}
\label{sec:CA}
Learning a mapping from text to color is inherently multimodal. For instance, a text `autumn' can be mapped to a variety of plausible color palettes. As text becomes longer, such as `midsummer to autumn' or `autumn breeze and falling leaves', the scope of possible matching palettes becomes more broad and diverse. To appropriately model the multimodality of our problem, we utilize the conditioning augmentation (CA)~\cite{zhang2017stackgan} technique. Rather than using the fixed sequence of encoded text as input to our generator, we randomly sample latent vector $\hat{c}$ from a Gaussian distribution $\mathcal{N}(\mu(h),\Sigma(h))$ as shown in Fig.~\ref{fig:ca_model}. This randomness allows our model to generate multiple plausible palettes given same text input. 

To obtain the conditioning variable $\hat{c}=\{\hat{c}_{1},\cdots,\hat{c}_{T}\}$,
the pre-trained word vectors $x=\{x_{1},\cdots,x_{T}\}$ are first fed into a GRU encoder to compute hidden states $h=\{h_{1},\cdots,h_{T}\}$. This text representation is fed into a fully-connected layer to generate $\mu$ and $\sigma$ (the values in the diagonal of $\Sigma$) for the Gaussian distribution $\mathcal{N}(\mu(h),\Sigma(h))$. Conditioning variable $\hat{c}$ is computed by $\hat{c}=\mu+\sigma\odot\epsilon$, where $\odot$ is the element-wise multiplication and $\epsilon\sim\mathcal{N}(0,I)$. The resulting set of vectors $\hat{c}=\{\hat{c_{1}},\cdots,\hat{c_{T}}\}$ will be used as \emph{condition} for our generator.

\paragraph{Generator.}
\label{sec:TPN Generator}
We design our generator $G_{0}$ as a variant of a GRU decoder with attention mechanism~\cite{luong2015effective,bahdanau2014neural,cho2014learning}. The $i$-th color of the palette $\hat{y}_{i}$ is computed as
\begin{equation}
\hat{y}_{i}=f(s_{i})\;\;\mathrm{where}\;\;s_{i}=g(\hat{y}_{i-1},c_{i}, s_{i-1}).
\end{equation}
$s_{i}$ is a GRU hidden state vector for time $i$, having the previously generated color $\hat{y}_{i-1}$, the context vector $c_{i}$, and the previous hidden state $s_{i-1}$ as input. The GRU hidden state $s_{i}$ is given as input to a fully-connected layer $f$ to output the $i$-th color of the palette $\hat{y}_{i}\in\mathbb{R}^{3}$. The resulting five colors are combined to produce a single palette output $\hat{y}$. 

The context vector $c_{i}$ depends on a sequence of conditioning
vectors $\hat{c}=\{\hat{c_{1}},\cdots,\hat{c_{T}}\}$ and the previous hidden state $s_{i-1}$. The context vector $c_{i}$ is computed as the weighted sum of these conditions $\hat{c_{i}}$'s, i.e., 
\begin{equation}
c_{i}=\sum_{j=1}^{T}\alpha_{ij}\hat{c}_{j}.
\end{equation}
The weight $\alpha_{ij}$ of each conditional variable $\hat{c}_{j}$ is computed by
\begin{equation}
\alpha_{ij}=\frac{\exp(e_{ij})}{\sum_{k=1}^{T}\exp(e_{ik})}
\;\;\mathrm{where}\;\;
e_{ij}=a\left(s_{i-1},\hat{c}_{j}\right).
\end{equation}
\begin{equation}
a\left(s_{i-1},\hat{c}_{j}\right)=w^T\sigma(W_{s}s_{i-1}+W_{\hat{c}}\hat{c}_{j}),
\end{equation}
where $\sigma(\cdot)$ is a sigmoid activation function and $w$ is a weight vector. The additive attention~\cite{bahdanau2014neural} $a\left(s_{i-1},\hat{c}_{j}\right)$ computes how well the $j$-th word of the text input matches the $i$-th color of the palette output. The score $\alpha_{ij}$ is computed based on the GRU hidden state $s_{i-1}$ and the $j$-th condition $\hat{c}_{j}$. The attention mechanism enables the model to effectively map complex text input to the palette output.

\paragraph{Discriminator.}
For the discriminator $D_{0}$, the conditioning variable $\bar{c}$ and the color palette are concatenated and fed into a series of fully-connected layers. By jointly learning features across the encoded text and palette, the discriminator classifies whether the palettes are real or fake.  

\subsection{Palette-based Colorization Networks (PCN)}
\label{sec:PCN}
\subsubsection{Objective Function}
The goal of the second networks is to automatically produce
colorizations of a grayscale image guided by the color palette as a conditioning variable. The inputs are a grayscale image $L\in\mathbb{\mathbb{\mathbb{R^{\mathit{H\times W\times1}}}}}$
representing the lightness in CIE \emph{Lab} space and a color palette
$p\in\mathbb{R^{\mathrm{15}}}$ consisting of five colors in \emph{Lab}
values. The output $\hat{I}\in\mathbb{R^{\mathit{H\times W\times2}}}$
corresponds to the predicted \emph{ab} color channels of the image. The
objective function of the second model can be expressed as
\begin{equation}
L_{D_{1}}=\mathbb{E_{\mathit{I\sim P_{data}}}}[\log D_{1}(p,I)]+\mathbb{E_{\mathrm{\mathit{\hat{I}\sim P_{G_{1}}}}}\mathrm{[\log(1-\mathit{D}_{1}(\mathit{p,\hat{I}}))],}}
\end{equation}
\begin{equation}
L_{G_{1}}=\mathbb{E_{\mathit{\hat{I}\sim P_{G_{1}}}}}[\log(1-\mathit{D}_{1}(\mathit{p,\hat{I}}))]+\lambda_{H}L_{H}(\hat{I},I).
\end{equation}
$D_{1}$ and $G_{1}$ included in the equation are shown in Fig.\ref{fig:full_model}. We have also added the Huber loss to the generator's objective function. In other words, the generator learns to be close to the ground truth image with \emph{plausible} colorizations, while incorporating palette colors to the output image to fool the discriminator. We set $\lambda_{H}=10$ in our model.
 
\subsubsection{Networks Architecture}
\label{sec:PCN_architecture}
\paragraph{Generator.}
The generator consists of two sub-networks: the main colorization networks and the conditioning networks. Our main colorization networks adopts the U-Net architecture~\cite{ronneberger2015u}, which has shown promising results in colorization tasks~\cite{isola2017image,zhang2017real}. The skip connections help recover spatial information~\cite{ronneberger2015u}, as the input and the output images share the location of prominent edges~\cite{isola2017image}. 

The role of the conditioning networks is to apply the palette colors to the generated image. During training, the networks are given a palette $p\in\mathbb{R}^{15}$ extracted from the ground truth image $I$. We utilize the Color Thief~\footnote{\url{http://lokeshdhakar.com/projects/color-thief/}} function to extract a palette consisting of five dominant colors of the ground truth image. Similar to the previous work~\cite{zhang2017real}, the conditioning palette $p$ is fed into a series of $1\times1$ \textit{conv-relu} layers as shown in Fig.~\ref{fig:full_model}. The feature maps in layers 1, 2, and 4 are duplicated spatially to match the spatial dimension of the $conv9$,
$conv8$, and $conv4$ features in the main colorization networks and added in an element-wise manner. The palette $p$ is fed into upsampling layers with skip connections as well as the middle of the main networks. This allows the generator to detect prominent edges and apply palette colors to suitable locations of the image. During test time, we use the generated palette $\hat{y}$ from the first networks (TPN) as the conditioning variable, colorizing the grayscale image with the predicted palette colors. 

\paragraph{Discriminator.}
As our discriminator $D_{1}$, we use a variant of the DCGAN architecture~\cite{radford2015unsupervised}. The image and conditioning variable $p$ are concatenated and fed into a series of \textit{conv-leaky relu} layers to jointly learn features across the image and the palette. Afterwards, it is fed into a fully-connected layer to classify whether the image is real or fake.

\subsection{Implementation Details}
We first train $D_{0}$ and $G_{0}$ of TPN for 500 epochs using the PAT dataset. We then train $D_{1}$ and $G_{1}$ of the PCN for 100 epochs, using the extracted palette from a ground truth image. Finally, we use the trained generators $G_{0}$ and $G_{1}$ during test time to colorize a grayscale image with generated palette $\hat{y}$ from a text input $x$. All networks are trained using Adam optimizer~\cite{kingma2014adam} with a learning rate of 0.0002. Weights were initialized from a Gaussian distribution with zero mean and standard deviation of 0.05. We set other hyper parameters as $\delta=1,\lambda_{H}=100$, and $\lambda_{KL}=0.5$.

\section{Experimental Results}
 
This section presents both quantitative and qualitative analyses of our proposed model. We evaluate the TPN (Section~\ref{sec:TPN}) based on our PAT dataset. For the training of the PCN (Section~\ref{sec:PCN}), we use two different datasets, CUB-200-2011 (CUB)~\cite{WahCUB_200_2011} and ImageNet ILSVRC Object Detection (ImageNet dataset)~\cite{russakovsky2015imagenet}. 

\begin{figure}[t]
\centering
\includegraphics[width=\textwidth]{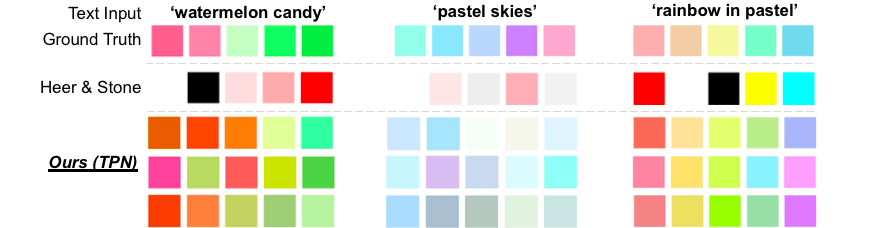}
\caption{\textbf{Comparison to baselines and qualitative analysis on multimodality:} Our TPN generates appealing color palettes that reflect all details of the text input. Also our model can generate multiple palettes with the same text input(three rows from bottom). In comparison, Heer and Stone~\cite{heer2012color}`s model frequently generates unrelated colors and has deterministic outputs.}\label{fig:multimodal}
\end{figure}

\subsection{Analysis on Multimodality and Diversity of Generated Palettes}
This section discusses the evaluation on multimodality and diversity of our generated palettes. Multimodality refers to how many different color palettes a single text input can be mapped to. In other words, if a single text can be expressed with more color palettes, the more multimodal it is. As shown in Fig.~\ref{fig:multimodal}, our model is multimodal, while previous approaches are deterministic, meaning that it generates only a particular color palette when given a text input. 
Diversity within a palette refers to how diverse the colors included in a single palette are. Following the current standard for perceptual color distance measurement, we use the CIEDE2000~\cite{sharma2005ciede2000} on CIE \emph{Lab} space to compute a model's multimodality and diversity. To measure multimodality, we compute the average minimum distances between colors from different palettes. To measure diversity of a color palette, we measure the average pairwise distance between the five colors within a palette. All measurements are computed based on the test dataset.

\paragraph{Results.}
Table~\ref{table:diversity} shows the multimodality and diversity measurement among the variants of our model. The CA module (Section~\ref{sec:CA}) enables our networks to suggest multiple color palettes when given the same text input. The model variant without CA (the first row in Table~\ref{table:diversity}) results in zero multimodality, indicating that the networks generate identical palettes for the same text input. Another palette generation model by Heer and Stone~\cite{heer2012color} also has zero multimodality. This shows that TPN is the only existing model that can adequately express multimodality, which is crucial in the domain of colors. Although Heer and Stone's model has higher diversity than TPN, Fig.~\ref{fig:multimodal} shows that their palettes contain irrelevant colors that may increase diversity but decrease palette quality. On the other hand, TPN creates those palettes containing colors that well match each other. Results on the fooling rate will be further illustrated in Section~\ref{sec:user_study}. 

\subsection{Analysis on Attention Outputs}
\begin{figure}[t]
\centering
\includegraphics[width=\textwidth]{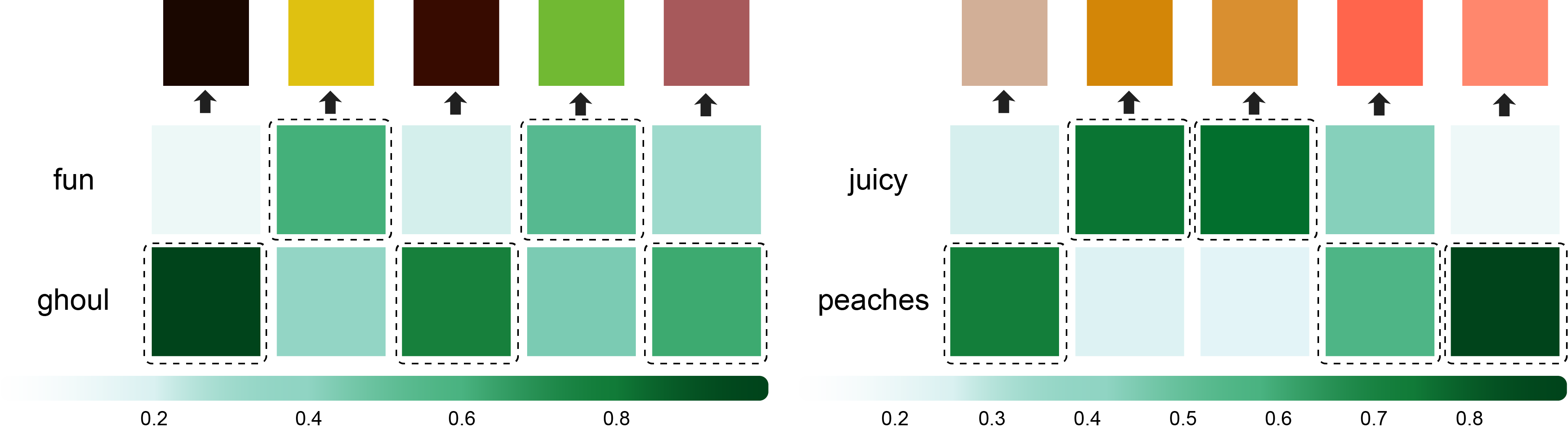}
\caption{\textbf{Attention analysis.} Attention scores measured by the TPN for two text input samples. Each box color (in green) denotes the attention score computed in producing the corresponding color shown on top. The dashed-line boxes indicate the word that each color output attended to.}\label{fig:attention}
\end{figure}

The attention module (Section~\ref{sec:TPN Generator}) plays a role of attending to particular words in text input to predict the most suitable colors for the text input. Fig.~\ref{fig:attention} illustrates how the predicted colors are influenced by attention scores. The green-colored boxes show attention scores computed for each word token when predicting each corresponding color in the palette. Higher scores are indicated by dashed-line boxes. We observe that three colors generated by attending to \textit{ghoul} are all dark and gloomy, while the other two colors attending to \textit{fun} are bright. This attention mechanism enables our model to thoroughly reflect the semantics included in text inputs of varying lengths.

\begin{figure}[t]
\centering
\includegraphics[width=\textwidth]{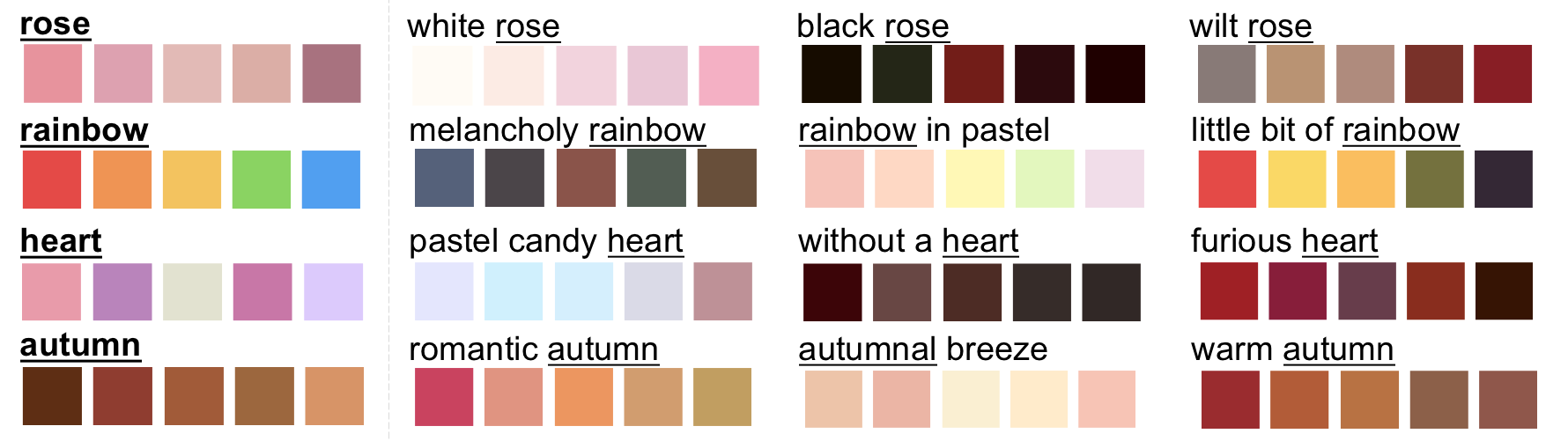}
\caption{\textbf{Qualitative analysis on semantic context.} Our model reflects subtle nuance differences in the semantic context of a given text input in the color palette outputs. Except for the first column, all the text combinations shown here are unseen data.}\label{fig:palette2}
\end{figure}

\begin{table}[t]
\begin{center}
\caption{\textbf{\label{table:diversity} Quantitative analysis results}}
\scalebox{0.85}{
\begin{tabular}{ccp{0.5cm}ccp{0.25cm}ccp{0.5cm}ccp{0.25cm}cc}
\toprule
&&& \multicolumn{5}{c}{\textbf{Palette Evaluation}} && \multicolumn{5}{l}{\textbf{User Study: Part I}}\\

\midrule
\\[-0.9em]

\multicolumn{2}{c}{\textbf{\ \ \ \ \ \ Model Variations}}&& 
\multicolumn{2}{l}{\textbf{Diversity}}  && \multicolumn{2}{l}{\textbf{Multimodality}}& & \multicolumn{5}{c}{\textbf{Fooling Rate (\%)}} \\ 
 
\\[-0.9em]
 Objective Function&CA&& Mean & Std && Mean & Std && Mean & Std && Max & Min  \\
 \\[-0.9em]
 \hline
 \\[-0.8em]
Ours (TPN)& X &&$19.36$&$8.74$ && $0.0$&$0.0$ && 
-&-&&-&-\\
Ours (TPN) & O && $20.82$  & $7.43$  &&$5.43$&$8.11$  &&
$\textbf{56.2}$  & $12.7$ && $\textbf{76.7}$&$37.1$\\
Heer and Stone & - &&$35.92$&$12.66$ && $0.0$&$0.0$ && $39.6$&$10.8$&&$58.2$&$25.8$ \\
 \hline

Ground truth palette & - && $32.60$  & $21.84$  &&-&- &&-&-&&-&-\\

 \hline
\end{tabular}}
\end{center}
\end{table}

\begin{figure}[t]
\centering
\includegraphics[width=\textwidth]{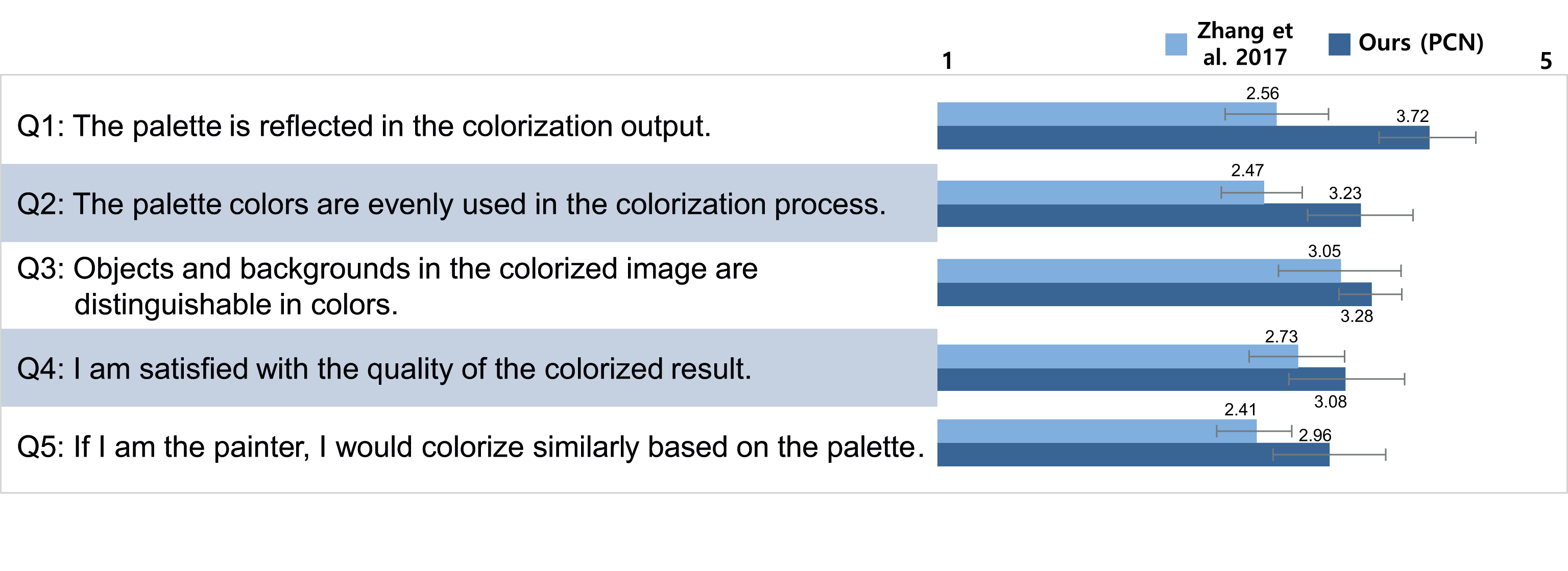}
\caption{\textbf{Colorization performance comparisons}. Mean and standard deviation values for each question are reported for the baseline~\cite{zhang2017real} and our PCN. Our PCN scores higher on all of the questions, showing that users are more satisfied with PCN.}\label{fig:user_study_result}
\end{figure}

\subsection{User Study} 
\label{sec:user_study}
We conduct a user study to reflect universal user opinions on the outputs of our model. Our user study is composed of two parts. The first part measures how the generated palettes match the text inputs. The second part is a survey that compares the performance of our palette-based colorization model to another state-of-the-art colorization model. 53 participants took part in our study.   

\subsubsection{Part I: Matching between Text and Generated Palettes}
Our goal is to generate a palette with a strong semantic connection with the given text input. A natural way to evaluate it is to quantify the degree of connection between the text input and the generated palette, in comparison to the same text input and its ground truth palette. Given a text input, its generated palette, and the ground truth palette, we ask human observers to select the palette that best suits the text input. A fooling rate (FR) in this study indicates the relative number of generated palettes chosen over ground truth palettes. More people choosing the generated palette results in a higher FR. This measure has often been used to assess the quality of colorization results~\cite{zhang2017real,guadarrama2017pixcolor}. We will use this metric to measure how much a text input matches its generated palette.

\paragraph{Study Procedure.}
Users participate in the user study over TPN and Heer and Stone's model~\cite{heer2012color}. Each consists of 30 evaluations. We randomly choose a single data item out of 992 test data and show the text input along with the generated palette and the ground truth palette.

\paragraph{Results.}
In Table~\ref{table:diversity}, we measure the FR score for each person and compute the mean and the standard deviation (std) of all of the scores from participants. Max and min scores represent the highest and the lowest FR scores, respectively, recorded by a single person. While Heer and Stone's model~\cite{heer2012color} shows low FR of 39.6\%, our TPN has the FR of 56.2\% while maintaining a high level of diversity and multimodality. The FR of 56.2\% indicates that the generated palettes are indistinguishable to human eyes and sometimes even match the input text better than the ground truth palettes. Note that the standard deviation of 12.7\% implies diverse responses to the same data pairs.        

\begin{figure}[t]
\centering
\includegraphics[width=\textwidth]{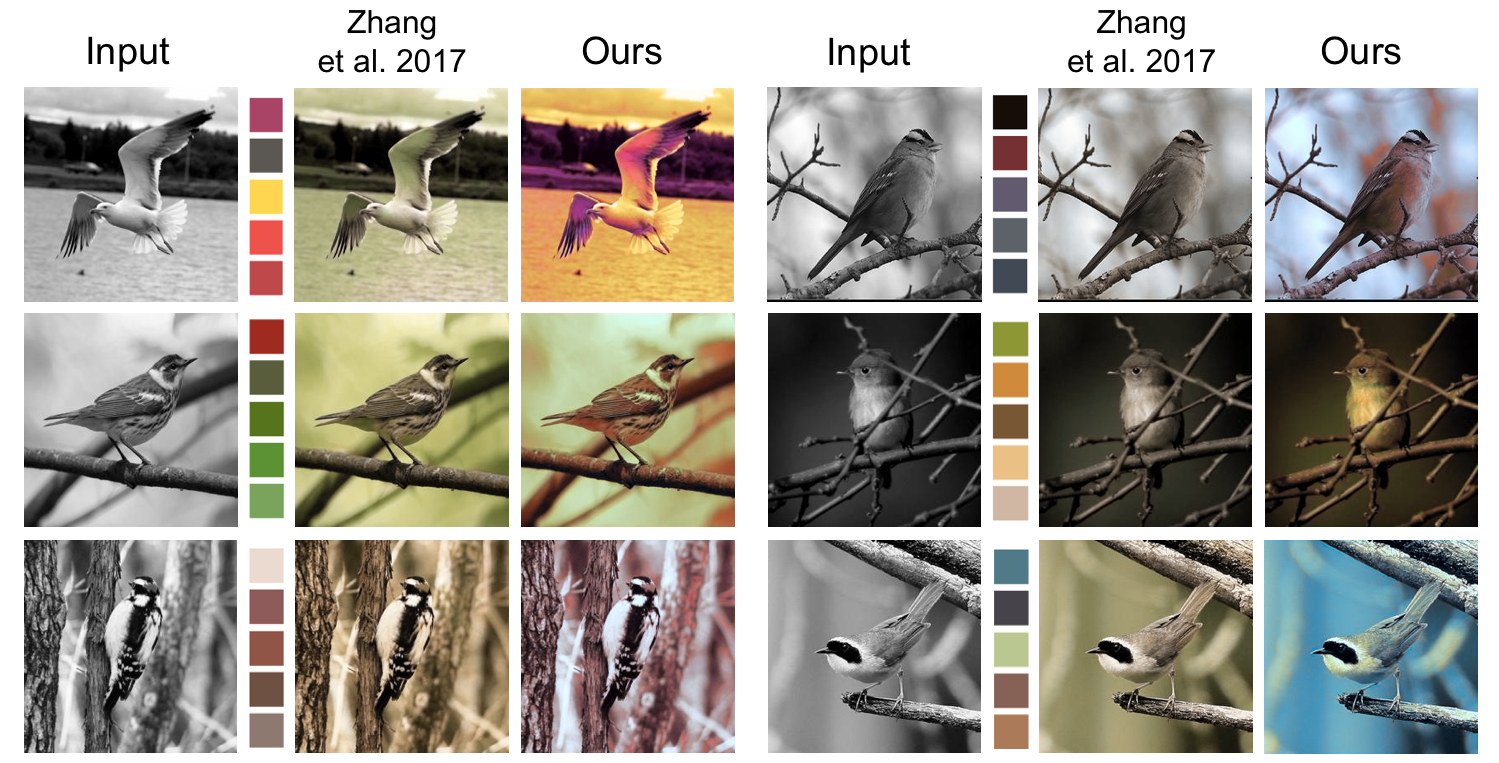}
\caption{We compare colorization results with previous work~\cite{zhang2017real}. The five-color palette used for colorization is shown next to the input grayscale image. Note that our PCN performs better at applying various colors included in the palette.}\label{fig:colorization_comparison}
\end{figure}

\subsubsection{Part II: Colorization Comparisons}
In this part of the user study, we conduct a survey on the performance of the PCN given palette inputs. Users are asked to answer five questions based on the given grayscale image, the color palette, and the colored image. For quantitative comparison, we set a state-of-the-art colorization model~\cite{zhang2017real} as our baseline. This model originally contains local and global hint networks. In our implementation of the baseline model, we utilize the global hint networks to infuse our generated palette to the main colorization networks. Note that we modified the baseline model to fit our task. Our novelty is the ability to produce high-quality colorization with only five colors of a palette while our baseline~\cite{zhang2017real} needs 313 bins of \emph{ab} gamut. Our model is able to colorize with limited information due to novel components such as the conditional adversarial loss and feeding the palette into skip-connection layers.

\paragraph{Study Procedure.} 
We show colorization results of our PCN and the baseline model one-by-one in a random order. Then, we ask each participant to answer five different questions (shown in Fig.~\ref{fig:user_study_result}) based on a five-point Likert scale. The focus of our questions is to evaluate how well the palette was used in colorizing the given grayscale image. The total number of data samples per test is 15.

\paragraph{Results.}
The resulting statistics are reported in Fig.~\ref{fig:user_study_result}. Our PCN achieves higher scores than the baseline model across all the questions. We can infer that the palettes generated by our model are preferred over palettes created by a human hand. Since our model learns consistent patterns from a large number of human-generated palette-text pairs, our model may have generated color palettes that more users could relate to.

\section{Conclusions}
We proposed a generative model that can produce multiple palettes from rich text input and colorize grayscale images using the generated palettes. Evaluation results confirm that our TPN can generate plausible color palettes from text input and can incorporate the multimodal nature of colors. Qualitative results on our PCN also show that the diverse colors in a palette are effectively reflected in the colorization results. Future work includes extending our model to a broader range of tasks requiring color recommendation and conducting the detailed analysis of our dataset.  

\vspace{0.1in}
\noindent\textbf{Acknowledgement.} This work was partially supported by the National Research Foundation of Korea (NRF) grant funded by the Korean government (MSIP) (No. NRF2016R1C1B2015924). Jaegul Choo is the corresponding author. 
\clearpage

%
%
%
%
\bibliographystyle{splncs04}
\bibliography{egbib}

\clearpage

\section*{Supplementary Materials}
\section{User Study Samples}
\begin{figure*}[h!]
\centering 
\includegraphics[width=0.99\textwidth]{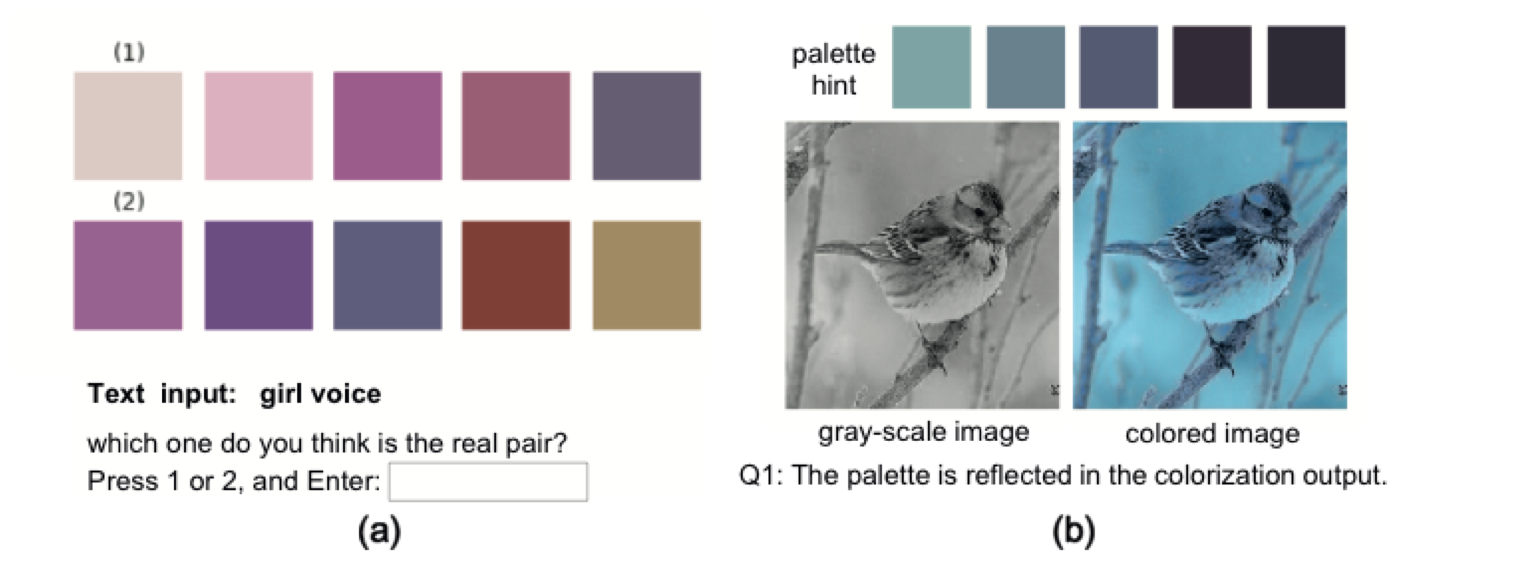}
\vspace*{-2mm}\caption{UI design of our user study.}\label{fig:user_study}
\end{figure*}

Our user study consists of two parts, one for evaluation of Text-to-palette Generation Networks (TPN) and the other for evaluation of Palette-based Colorization Networks (PCN). Fig.~\ref{fig:user_study}\textbf{(a)-(b)} illustrates how our data tuples were shown to the participants in \textbf{Part I} and \textbf{Part II}, respectively.

\section{Text-to-Palette Generation Networks (TPN)}
\subsection{Model Comparisons for Learning Global Color Distributions}
\begin{figure*}[h!]
\centering 
\includegraphics[width=0.99\textwidth]{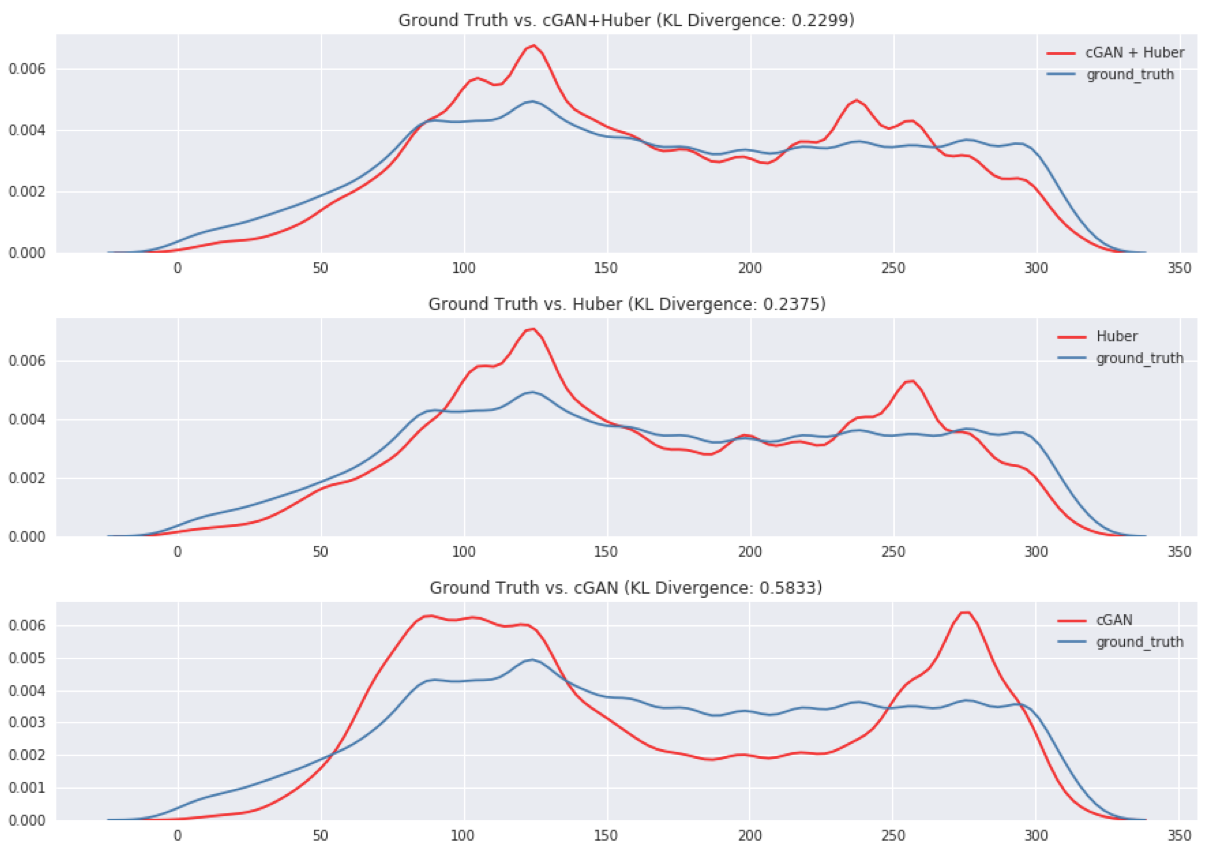}
\vspace*{-2mm}\caption{\textbf{Color Distribution Comparisons.} Red lines correspond to color distributions of generated palettes from three model variants. The blue lines denote the ground truth color distribution of the training data. The KL divergence of the three distribution pairs are computed as 0.2299, 0.2375, and 0.5833 in order.}\label{fig:Lab_distribution}
\end{figure*}

Fig.~\ref{fig:Lab_distribution} shows comparisons of color distributions between ground truth palettes of the training data and generated palettes from our test data. For each color distribution, we quantize the \emph{ab} values of every palette color into 313 color bins~\cite{zhang2016colorful} and visualize the probability distribution of \emph{ab} values. We compare three model variants of different objective functions: cGAN+Huber ($\lambda_{H}$=100), Huber ($\lambda_{H}$=100), and cGAN ($\lambda_{H}$=0). We also compute the Kullback-Leibler (KL) divergence between the ground truth palette distribution of the training data and that of our model variants. 

\begin{figure*}[h!]
\centering 
\includegraphics[width=0.99\textwidth]{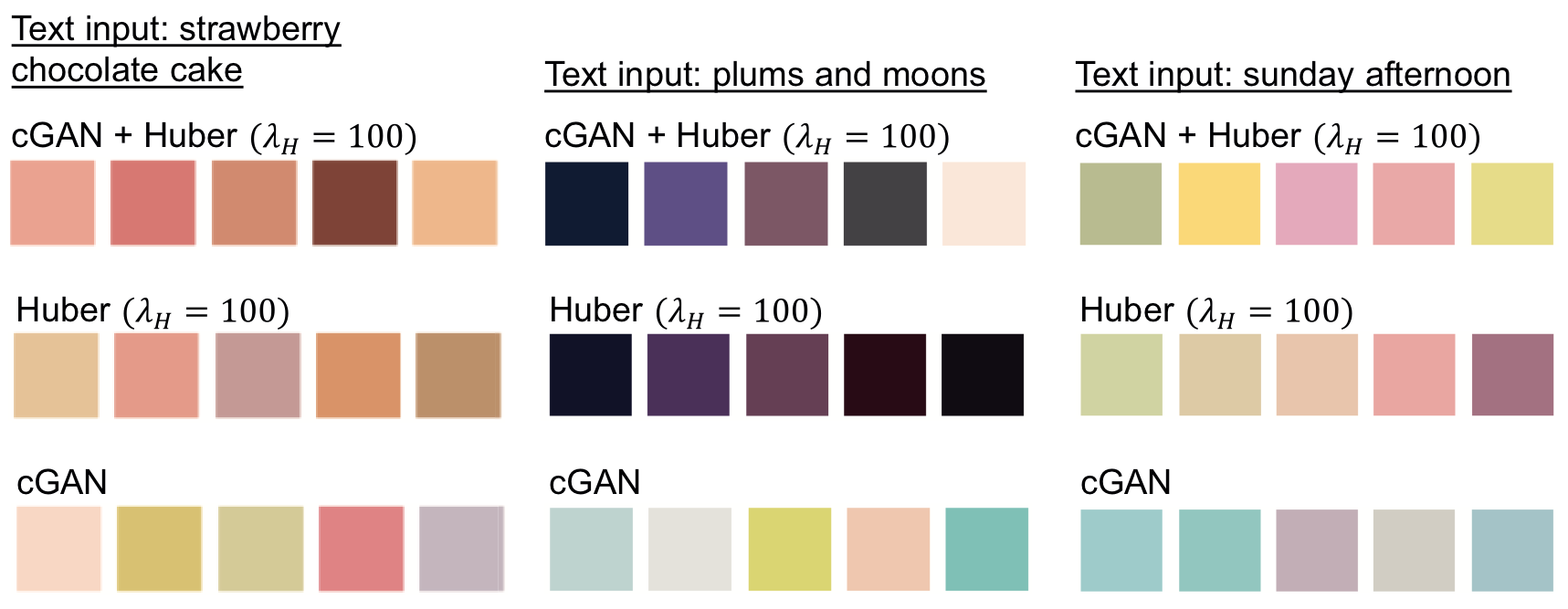}
\vspace*{-2mm}\caption{Comparison of palette prediction results from different model variations.}\label{fig:dist_palette_ex}
\end{figure*}

As shown in the bottommost plot of Fig.~\ref{fig:Lab_distribution}, the Huber loss plays a critical role in producing proper colors close to the ground truth image. Without the Huber loss, the model does not only fail to recover the color distribution similar to the ground truth data but also exhibits the lowest fooling rate of 30.7\% in user study results. On the other hand, the model with cGAN+Huber loss ($\lambda_{H}$=100) records the lowest KL divergence of 0.2299 as well as the best fooling rate of 56.2\%, while the model with only the Huber loss ($\lambda_{H}$=100) records the second best. This is due to the fact that only using the Huber loss leads to blindly averaging over multiple ground truth palettes, resulting in slightly desaturated palette results as shown in the second row of Fig.~\ref{fig:dist_palette_ex}. In contrast, the model with both cGAN+Huber loss learns and preserves various ground truth colors rather than simply averaging them, resulting in bright, highly saturated results as shown in the first row of Fig.~\ref{fig:dist_palette_ex}.

\subsection{Additional Results}
This section shows additional, diverse and detailed results from TPN.  
\begin{figure*}[h!]
\centering 
\includegraphics[width=0.99\textwidth]{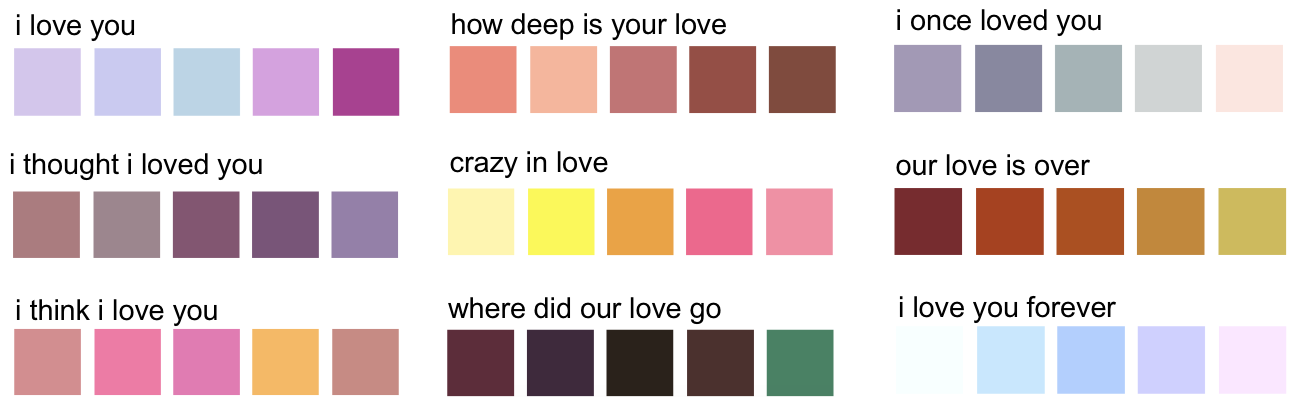}
\vspace*{-2mm}\caption{Handling phrase-level inputs about `love'.}\label{fig:love_variations}
\end{figure*}

Fig.~\ref{fig:love_variations} shows how our model handles phrase-level inputs. To make comparison easier, all the phrases contains the word `love.' It is interesting to see how our model chooses to express the subtle nuance differences included in the input text. Notice how the output color palettes tend to be darker for text inputs that are negative towards `love' (e.g., `i thought i loved you' and `where did our love go'). All input phrases included in this figure are unseen data.
\begin{figure*}[h!]
\centering 
\includegraphics[width=0.99\textwidth]{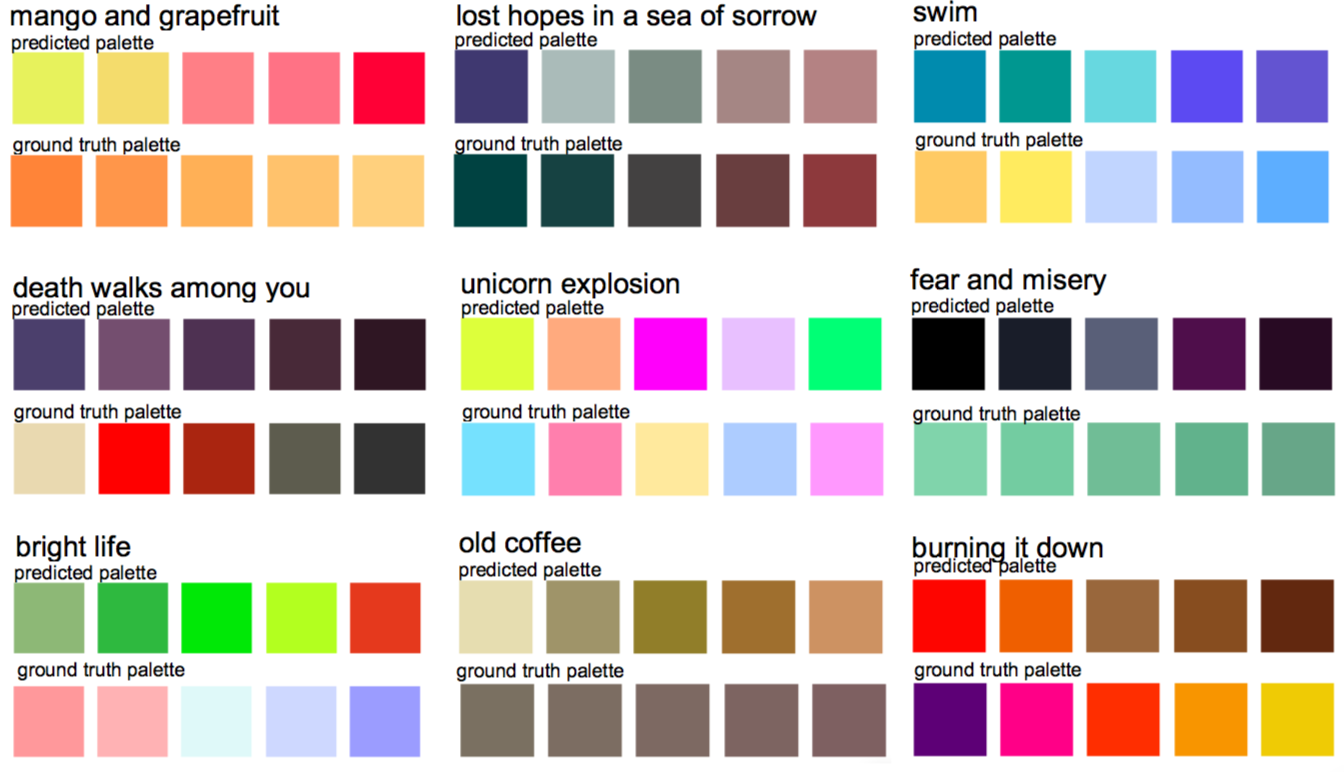}
\vspace*{-2mm}\caption{Palette predictions from test set.}\label{fig:test_palettes}
\end{figure*}

Fig.~\ref{fig:test_palettes} shows outputs of our model in comparison to ground truth palettes. If an input word is seen at least once in the training data, our model is able to output a color palette related to the input word. For instance, take a look at the color palette named `mango and grapefruit' on the top left. The word `grapefruit' is included only once in the training set. Yet, the model successfully outputs a color palette that matches the text input. Also, ground truth palettes are included for a direct comparison with generated palettes. Even if the predicted palette is not exactly identical to the ground truth palette, both can be perceived as reasonable colors.
\begin{figure*}[h!]
\centering 
\includegraphics[width=0.99\textwidth]{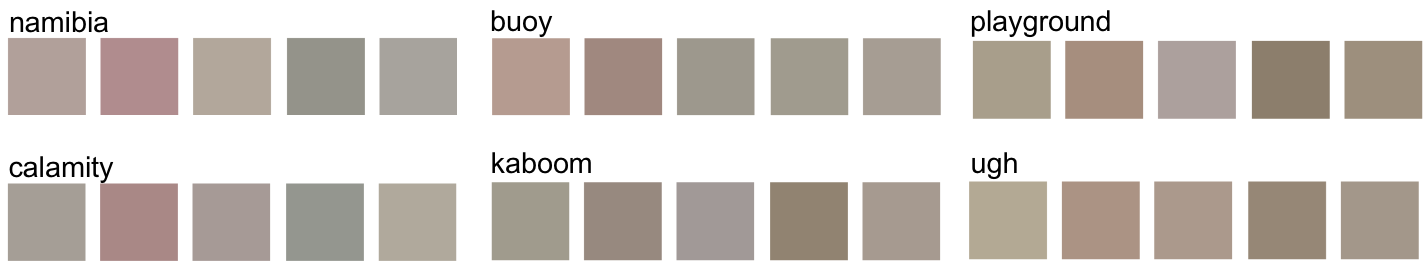}
\vspace*{-2mm}\caption{Failed results of TPN. Our model fails and outputs the same washed-out grayish-brown color palettes for unknown tokens.}\label{fig:failure_cases}
\end{figure*} 
Even though our model can effectively produce semantically meaningful colorizations, it struggles when unknown tokens are given as input. Unknown tokens refer to words not included in the training set. it is not surprising that our model fails and outputs the same washed-out grayish-brown palettes as we can see in Fig.~\ref{fig:failure_cases}. On the other hand, our model can still produce reasonable palettes in the case of unseen, new combinations of words found in the training set. For example, `bright life' in Fig.~\ref{fig:test_palettes} was seen separately as `bright' and `life' in the training set but not together. Thus, `bright life' is classified as unseen data, which our model has no problem in predicting color palettes from.

\section{Palette-Based Colorization Networks (PCN)}
We present additional colorization results on datasets including CUB-200-2011 (CUB dataset)~\cite{WahCUB_200_2011}, ImageNet ILSVRC Object Detection (ImageNet dataset)~\cite{russakovsky2015imagenet}, and Graphical Pattern images (Pattern images) in Figs.~\ref{fig:bird1}-\ref{fig:pattern2}. In these figures, the leftmost columns are grayscale images. Text inputs are given above the grayscale image. The vertical color palettes next to the grayscale images are palettes generated from the text input. The output has been colorized with the generated color palette. We would like to emphasize that our model effectively utilizes the generated color palettes during the colorization process. The colorized image may be different from its natural colors because our networks incorporate additional color hints. We display the original ground truth image on the right to compare how different an image becomes after applying the palettes.  
 
\subsection{CUB-200-2011}
\begin{figure*}[h!]
\centering 
\includegraphics[width=0.99\textwidth]{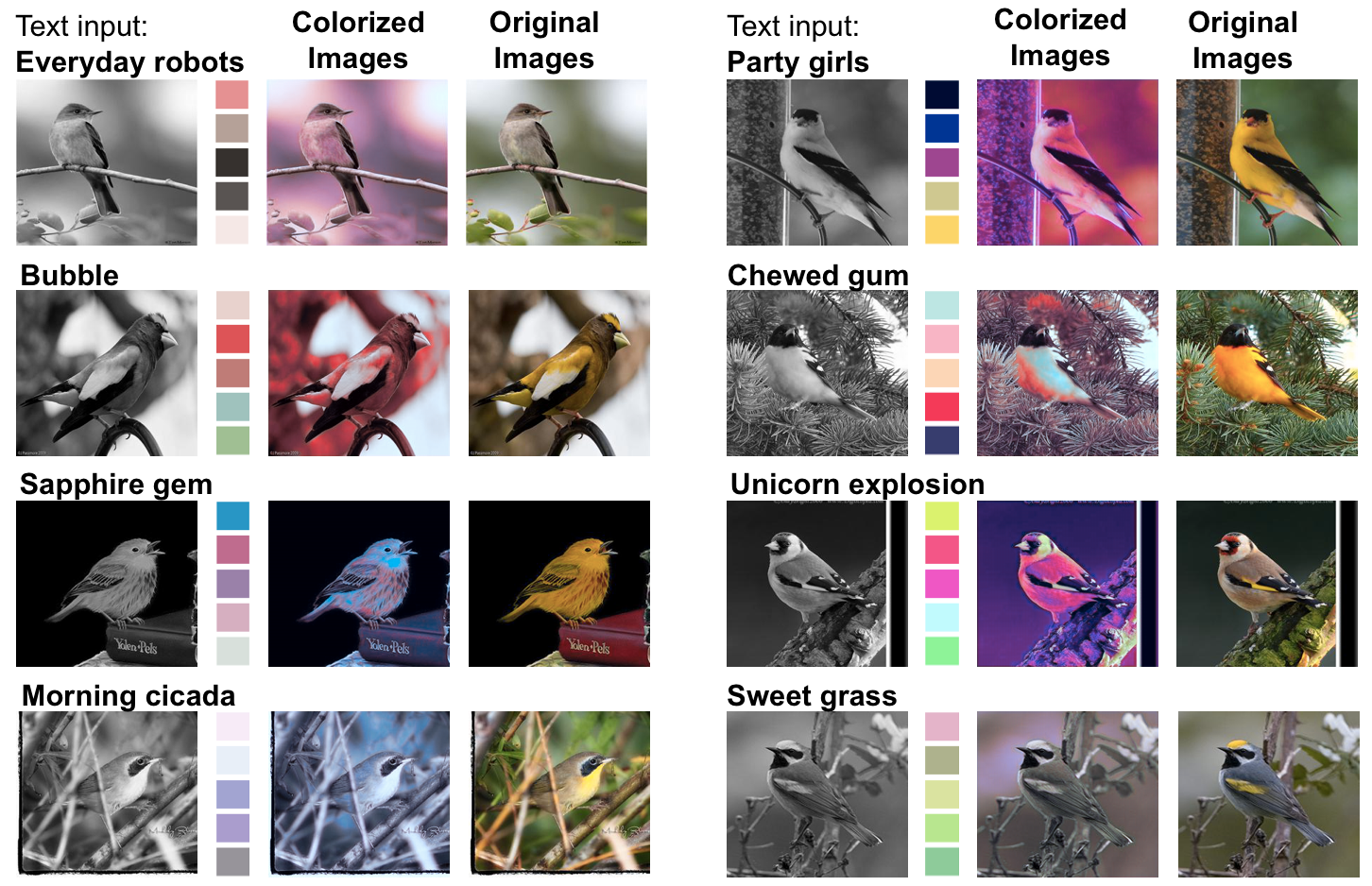}
\vspace*{-2mm}\caption{Results on CUB dataset (1).}\label{fig:bird1}
\end{figure*}

\begin{figure*}[h!]
\centering 
\includegraphics[width=0.99\textwidth]{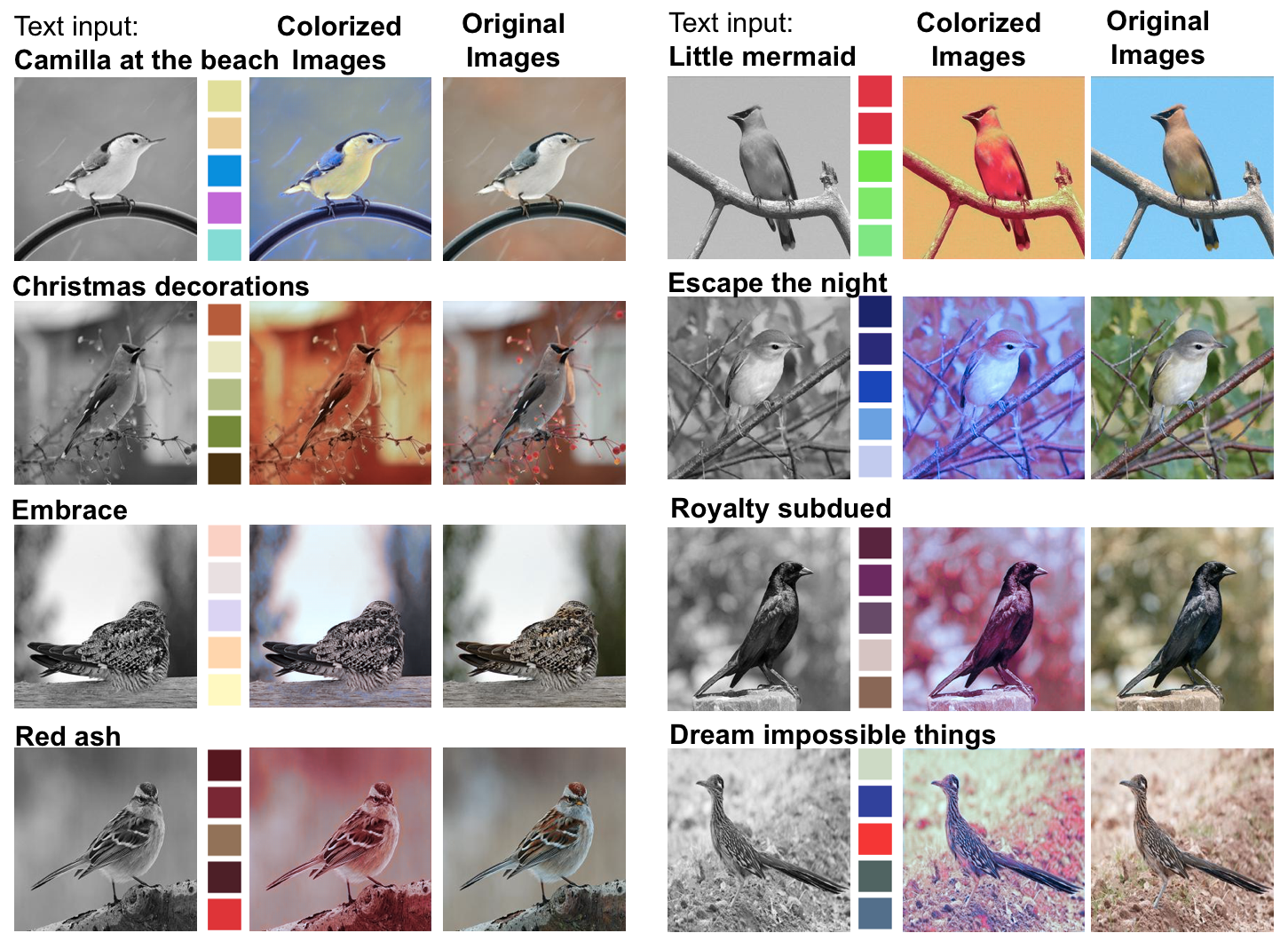}
\vspace*{-2mm}\caption{Results on  CUB dataset (2).}\label{fig:bird2}
\end{figure*}

Figs.~\ref{fig:bird1} and \ref{fig:bird2} show additional colorization results on the CUB dataset.

\subsection{ImageNet ILSVRC Object Detection}
\begin{figure*}[h!]
\centering 
\includegraphics[width=0.99\textwidth]{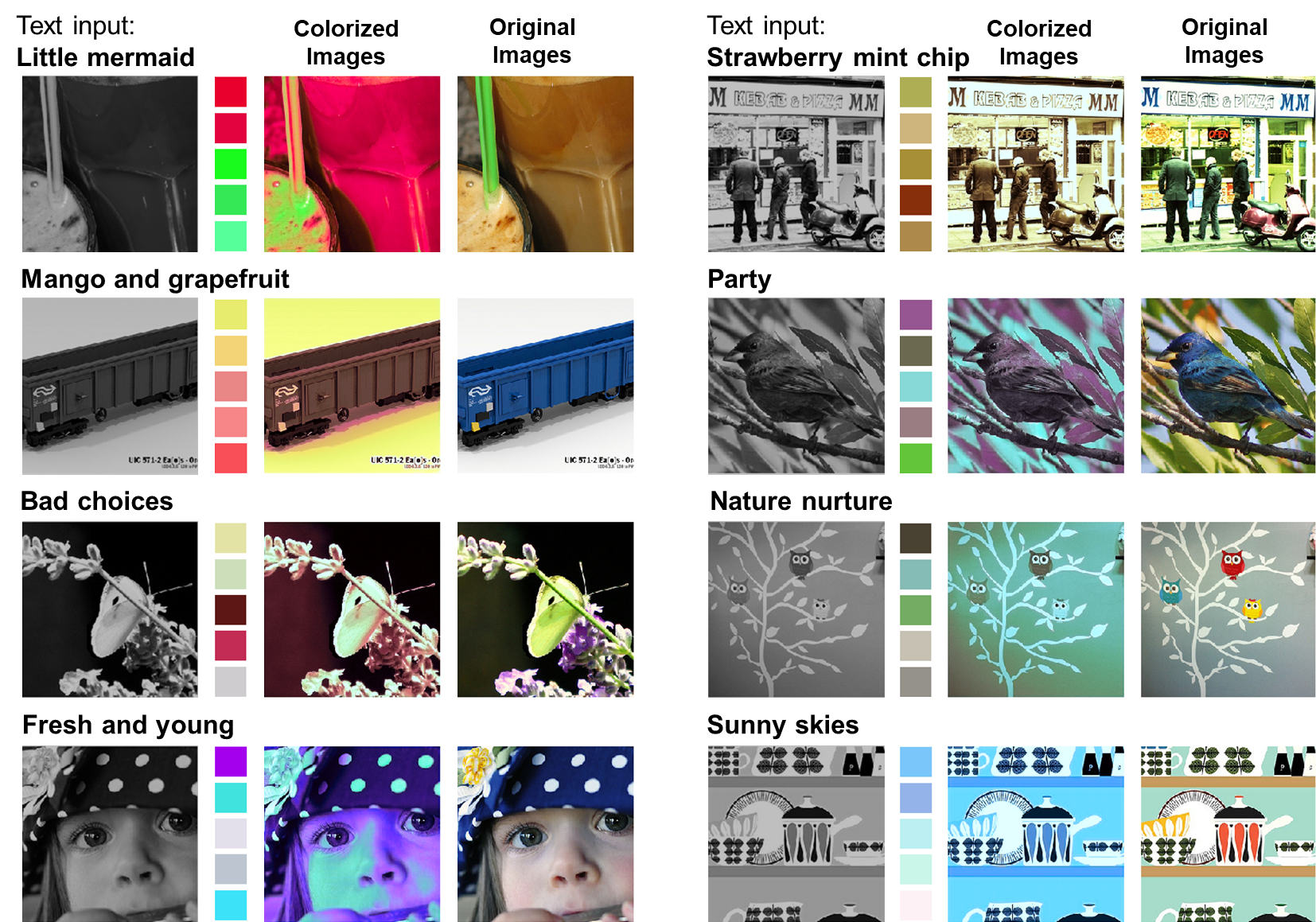}
\vspace*{-2mm}\caption{Results on ImageNet dataset (1).}\label{fig:imagenet1}
\end{figure*}

\begin{figure*}[h!]
\centering 
\includegraphics[width=0.99\textwidth]{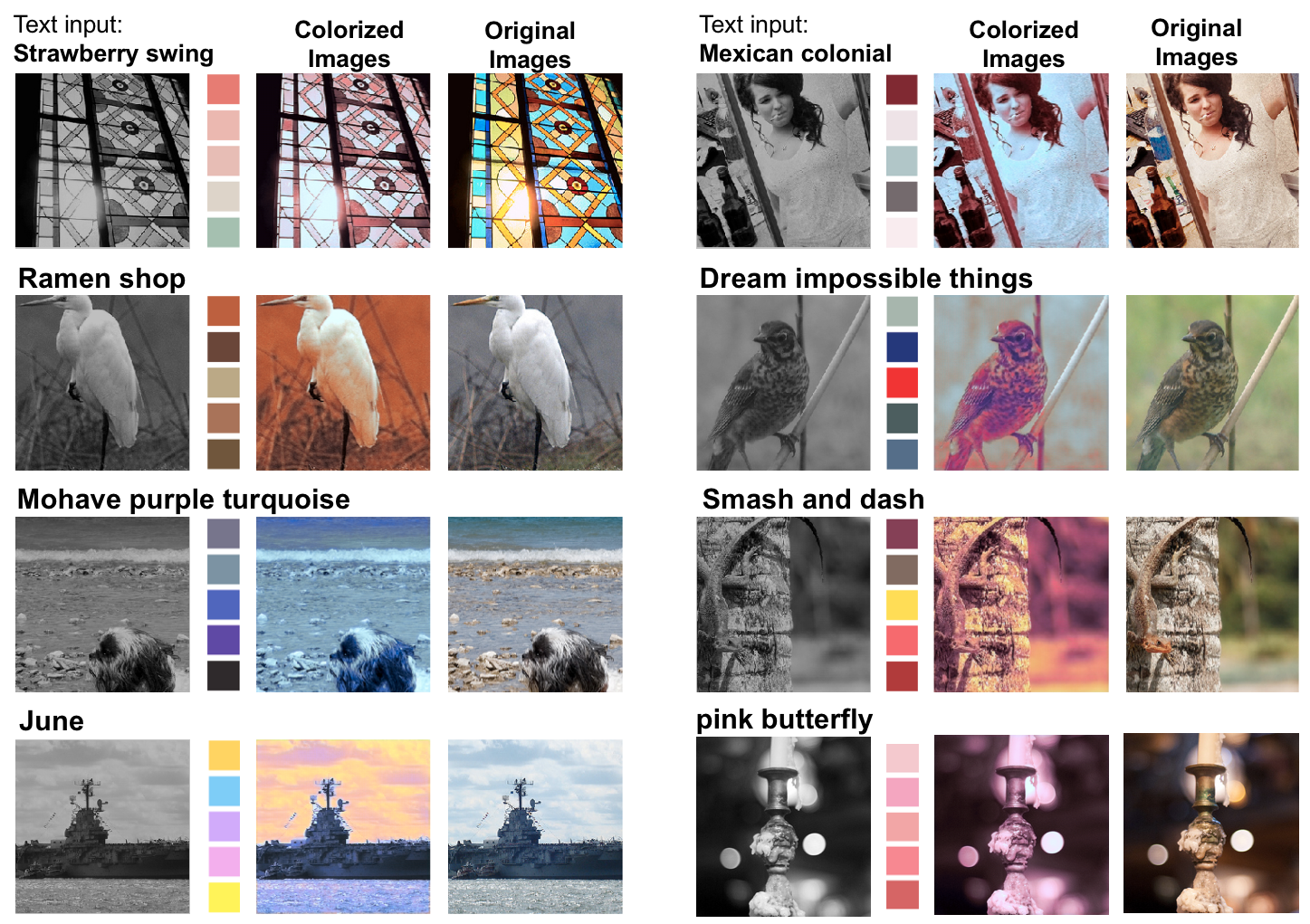}
\vspace*{-2mm}\caption{Results on ImageNet dataset (2).}\label{fig:imagenet2}
\end{figure*}

Figs.~\ref{fig:imagenet1} and \ref{fig:imagenet2} show additional colorization results on the ImageNet dataset.

\subsection{Graphical Pattern Images}
\begin{figure*}[h!]
\centering 
\includegraphics[width=0.99\textwidth]{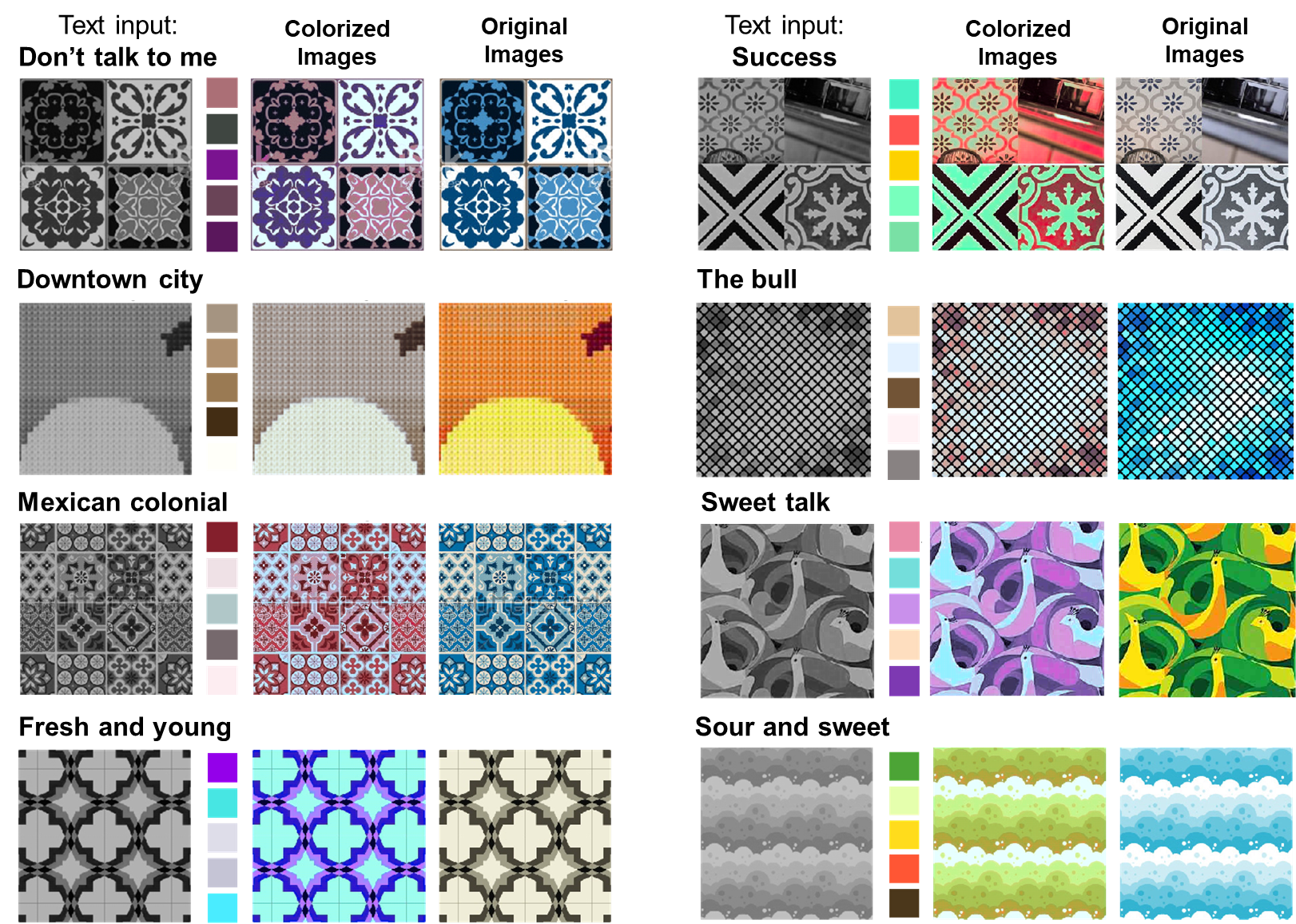}
\vspace*{-2mm}\caption{Results on graphical pattern images (1).}\label{fig:pattern1}
\end{figure*}

\begin{figure*}[t!]
\centering 
\includegraphics[width=0.99\textwidth]{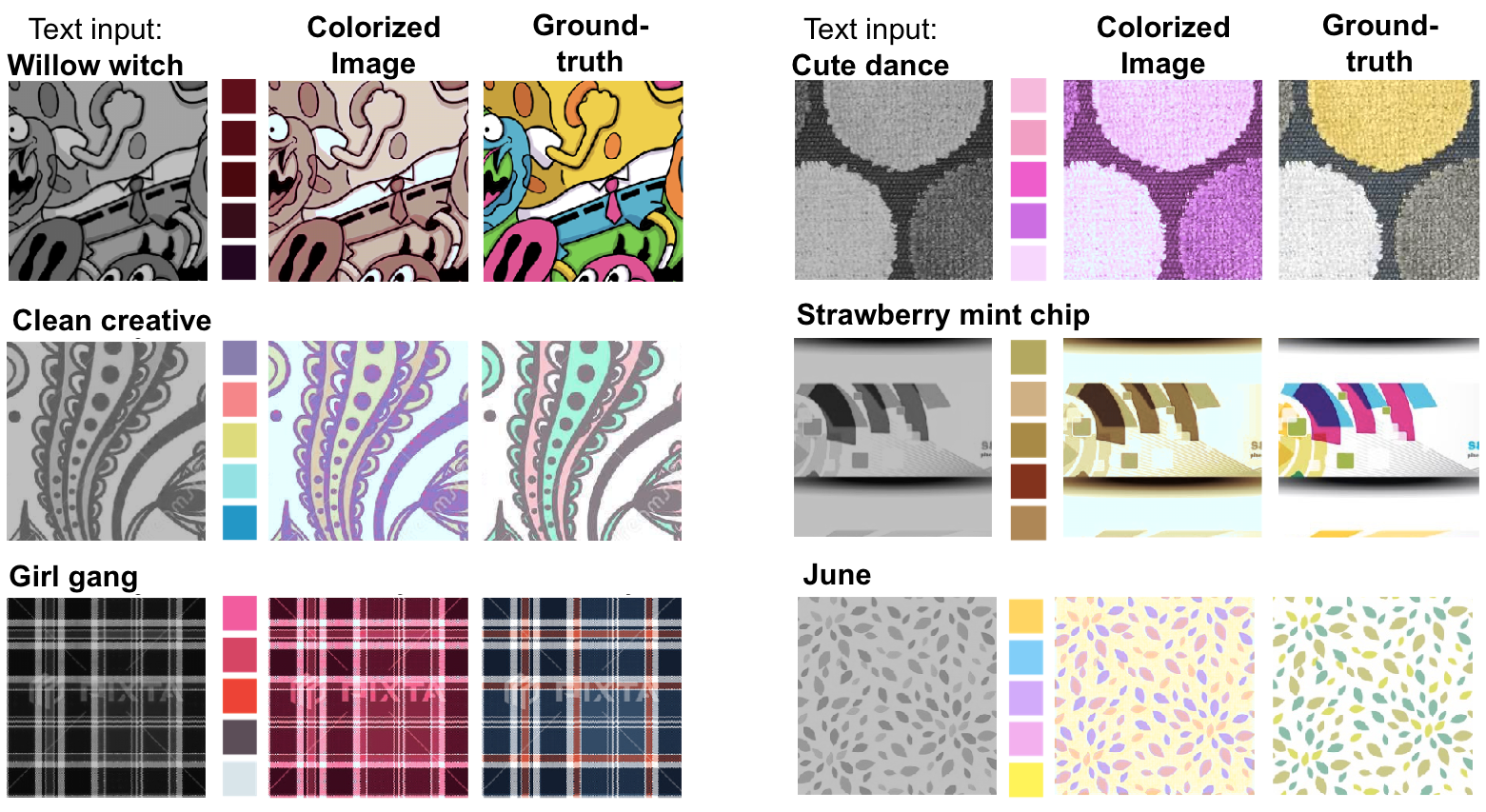}
\vspace*{-2mm}\caption{Results on graphical pattern images (2).}\label{fig:pattern2}
\vspace*{5in}
\end{figure*}

Our PCN model generalizes surprisingly well on other types of images. Our model is trained on ImageNet dataset, which is mostly made up of natural images. Instead of natural images, we used our colorization model to colorize graphical pattern images. The graphical pattern images are crawled from Google through searching keywords such as `pattern,' `fabric pattern,' or `beautiful patterns.' As seen in Figs.~\ref{fig:pattern1} and \ref{fig:pattern2}, graphical pattern images are significantly different from natural images. The colorized outputs show that our model can apply our generated color palettes on images of diverse shapes and textures. The results qualitatively show that our palette-based colorization model is transferable to other image domains. 

\clearpage

\end{document}